\documentclass[letterpaper, 10 pt, conference]{ieeeconf}  

\IEEEoverridecommandlockouts                              

\usepackage{cite}

\usepackage{amsmath}
\usepackage{amsfonts} 
\usepackage{graphicx}
\usepackage{hyperref}
\usepackage{subfig}

\renewcommand{\baselinestretch}{0.98}

\title{\LARGE \bf
Safe haptic teleoperations of admittance controlled robots with virtualization of the force feedback
}

\author{Lorenzo Pagliara\textsuperscript{1}, Enrico Ferrentino\textsuperscript{1}, Andrea Chiacchio\textsuperscript{2} and Giovanni Russo\textsuperscript{1}
\thanks{\textsuperscript{1} Department of Information Engineering, Electrical Engineering and Applied Mathematics (DIEM), University of Salerno
84084 Fisciano, Italy,
e-mail: \{lpagliara, eferrentino, giovarusso\}@unisa.it}
\thanks{\textsuperscript{2} Department of Medicine, Surgery and Dentistry (DIPMED), University of Salerno
84084 Fisciano, Italy,
e-mail:a.chiacchio998@gmail.com}
}

\begin{document}

\maketitle
\thispagestyle{empty}
\pagestyle{empty}

\begin{abstract}
Haptic teleoperations play a key role in extending human capabilities to perform complex tasks remotely, employing a robotic system. The impact of haptics is far-reaching and can improve the sensory awareness and motor accuracy of the operator. In this context, a key challenge is attaining a natural, stable and safe haptic human-robot interaction. Achieving these conflicting requirements is particularly crucial for complex procedures, e.g.\ medical ones. To address this challenge, in this work we develop a novel haptic bilateral teleoperation system (HBTS), featuring a virtualized force feedback, based on the motion error generated by an admittance controlled robot. 
This approach allows decoupling the force rendering system from the control of the interaction: the rendered force is assigned with the desired dynamics, while the admittance control parameters are separately tuned to maximize interaction performance. Furthermore, recognizing the necessity to limit the forces exerted by the robot on the environment, to ensure a safe interaction, we embed a saturation strategy of the motion references provided by the haptic device to admittance control. We validate the different aspects of the proposed HBTS, through a teleoperated blackboard writing experiment, against two other architectures. The results indicate that the proposed HBTS improves the naturalness of teleoperation, as well as safety and accuracy of the interaction.
\end{abstract}

\section{Introduction}
Robots are becoming increasingly employed in supporting humans in everyday activities. Their strength lies in their ability to perform tasks with extreme precision, repeatability and efficiency, substituting humans in situations that can be tedious or hazardous.
Although in most cases robots can operate autonomously, the high variability and complexity of some tasks require the expertise and adaptive skills of humans, combined with the robot's skills. In dental implant surgery, passive and semi-active robotic systems provide increased accuracy of the implant site position, reduced angular deviation and surgical time, besides increased success rate of the procedure \cite{yang_accuracy_2024}. Particularly, haptic-guided robotic systems allow the tracking of patient movements and the enforcement of motion constraints to the surgeon, to avoid errors with respect to the surgical plan \cite{bolding_accuracy_2022}. Besides passive and semi-active robots, in medical procedures, telerobots are also particularly popular and widespread.

\textit{Teleoperation}, defined as the operation of a system located remotely from the user \cite{lichiardopol_survey_2007}, is used in various fields, such as search and rescue operations, telesurgery, and space exploration, as it overcomes barriers in terms of distance and hazardous environments, and scales beyond human reach. The aim of a teleoperation system is to let the human operator experience \textit{telepresence} \cite{raju_design_1989}. This is only achievable if provided with enough sensory information, i.e. visual, auditory, and haptic, displayed in a natural way to give the illusion of being immersed in the remote environment. Since the sense of touch is a central aspect of human manipulation in contact-rich tasks, \textit{haptic teleoperations} play a key role in achieving successful and safe interaction. Generally, the typologies of haptic feedback provided in a HBTS are based on the two complementary modalities of the human sense of touch \cite{el_rassi_review_2020}: \textit{kinaesthetic}, including forces and torques, which enable the perception of the body's position in space, relative to other objects; \textit{tactile}, which enables the perception of shape, temperature, vibration, stiffness, and texture. Force feedback has historically been the focus of haptic feedback research, as it increases the \textit{transparency} of teleoperation, defined as the ability to perceive the remote environment as directly encountered \cite{hirche_human_2007}. Such feedback can potentially introduce oscillations into the control loop, causing instability of the overall system. On the other hand, the tactile feedback does not affect the \textit{stability}, but greatly reduces the transparency of teleoperation, as it does not allow a natural and accurate transfer of the sensed forces and the impedance of the remote environment to the human operator. Therefore, stability and transparency are two conflicting factors that require a suitable trade-off to ensure accuracy and effectiveness of an HBTS.

Furthermore, interaction scenarios require the robot to exhibit a compliant and adaptable behavior to ensure safe interaction with the surroundings. Such a behavior is achieved through force perception and control, as well as through collision avoidance and adaptive control techniques. Typically, impedance control \cite{hogan_impedance_1984} is employed to assign mass-spring-damper dynamics to the end-effector, or admittance control for position-controlled robots. These two controllers, in their standard design, cannot guarantee stringent safety requirements, such as those required in the certification process of applications with pHRI \cite{vavra_tips_2016}. In particular, ISO/TS 15066 defines limit values for contact forces with different parts of a human operator's body.

As for identifying a suitable trade-off in an HBTS, replacing force feedback with different sensory modalities is an approach to solve instability issues. Such approach is known as sensory substitution, and involves the use of auditory stimuli or visual displays \cite{kitagawa_effect_2004, cesari_sensory_2023}. The da Vinci Surgical System\textsuperscript{TM}, for instance, exploits the visual force feedback to avoid tissue damage during surgical procedures \cite{reiley_effects_2008}. The alternative approach is to combine the different feedback typologies into a single multi-modal platform, so as to offer an experience as close as possible to the actual human sense of touch \cite{abiri_multi-modal_2019}. Although some of the research on these approaches provide some promising solutions based on platforms that combine different feedback modalities, the main interest is in force feedback, as it provides the most natural transfer of the remote interaction to the human operator. Indeed, in order to let the human operator perceive contact forces, most works in the literature focus on force/torque sensor measurement-based force feedback \cite{panzirsch_exploring_2022, gong_improved_2022, dekker_design_2023}. As discussed in \cite{ji_improving_2023}, humans are able to perceive both interaction events, i.e. contact state transitions, collision and sliding, and effects, i.e. contact forces, reaction and friction. The lack of perception of any one of these may compromise the teleoperation operability. Force/torque measurement-based haptic feedback does not allow for a clear distinction of interaction events, which causes the human operator to be pushed back strongly due to the forces exerted by the robot being greater than expected. To address this issue, \cite{ji_improving_2023} introduces a haptic feedback pipeline capable of rendering interaction forces distinctly for each contact state. To mitigate sudden changes in the forces, \cite{ryu_passive_2010} introduces a virtual mass-spring system acting as a low-pass filter. 

As for safety requirements in interaction scenarios, building precisely on the certification process of applications with pHRI, \cite{lachner_energy_2021} proposes a robot energy control approach to limit the energy of the system below a safety budget; \cite{gao_combined_2023} proposes a combined active and passive variable admittance control architecture, to mitigate collisions and ensure the application of the desired force.

In this work we address the problem of improving the execution of precision interaction tasks via haptic teleoperations, by designing an HBTS which aims to fulfill more stringent safety requirements and to improve both the stability of the teleoperation and the naturalness for the human operator.
This is achieved by (1) making the robot behave compliantly with the remote environment, both explicitly and implicitly: explicitly in that it is subject to admittance control, implicitly in that interaction forces are rendered on the haptic device, allowing the human operator to react to them; (2) limiting the interaction forces between the robot and the environment by designing a saturation strategy of the motion references provided to the robot controller; (3) virtualizing the force feedback, despite the presence of a force/torque sensor, through the motion error generated by admittance control, which is input to a virtual spring-damper system.

We validate the proposed HBTS against two other architectures, including one based on force/torque sensor measurement-based force feedback, according to both quantitative and qualitative evaluation metrics, on a contact-rich task, consisting of blackboard writing with chalk.

\section{Methodology}
\subsection{System design}
\begin{figure}
\centering
\includegraphics[width=\columnwidth]{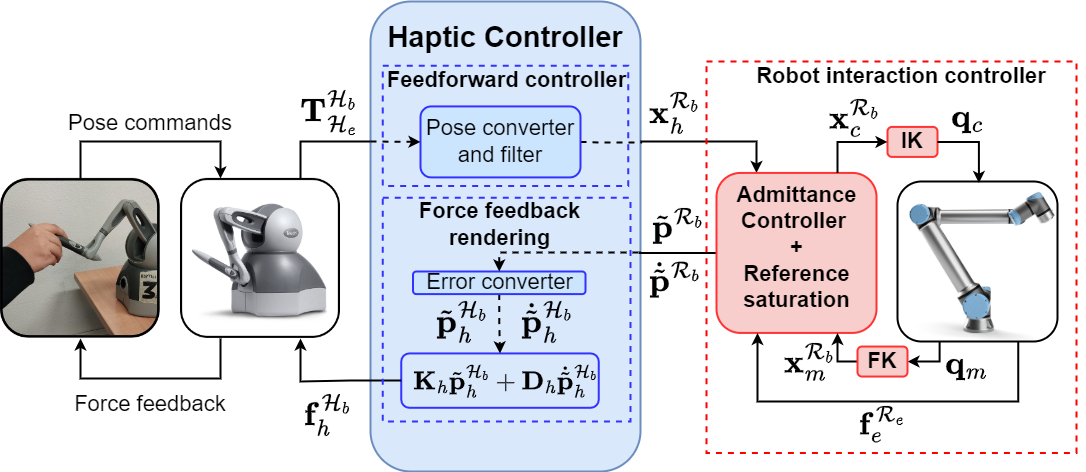}
\caption{Haptic bilateral teleoperation system scheme.}
\label{fig:haptic-bilateral-teleoperation-system}
\vspace{-7mm}
\end{figure}

We propose and validate a HBTS architecture, shown in Figure \ref{fig:haptic-bilateral-teleoperation-system}, including a master control station and a slave robot (SR). The SR is a robotic arm equipped with a 6-axis force/torque sensor: it is responsible of measuring the contact forces and supporting the tool that physically interacts with the environment. 
The SR is subject to admittance control to ensure stability and safety of teleoperation during the interaction phases, preventing damage to the arm or to the remote environment. Such a controller features the saturation of the motion references, as well as gravity compensation of the tool's payload. The objective of saturation and its operation are explained in Section \ref{sec:interaction-force-limitation-strategy}.

The master control station features a haptic device (HD) and a haptic controller (HC): it is responsible of generating the SR's motion references, and rendering the haptic feedback to let the human operator perceive the remote interaction forces. In addition, to generate stable references for the SR, the HC features the filtering of human tremors. The generation of motion references and their filtering are explained in Section \ref{sec:feedforward-haptic-controller}. The haptic feedback rendering methodology is explained in Section \ref{sec:force-feedback-rendering}.

\subsection{Feedforward haptic controller} \label{sec:feedforward-haptic-controller}
One of the main features of the HC is capturing the HD stylus poses with respect to the HD base frame $\mathcal{H}_b$, and transforming them into robot end-effector poses with respect to the base reference frame $\mathcal{R}_b$, so that the robot follows the movements of the HD stylus. This type of linear mapping, in which poses are simply transferred from HD space to robot space, is defined 1:1 position mapping \cite{radi_workspace_2012}, and is particularly suitable for the considered task, as it provides the right trade-off between the speed of movements and the workspace size.

We denote with ${\bf T}_{b}^{a} \in SE(3)$ the homogeneous transformation matrix from frame $a$ to frame $b$, in the form
\begin{equation} \label{eq:transform-matrix}
    {\bf T}_{b}^{a} =
    \begin{pmatrix}
        {\bf R}_{b}^{a} & {\bf p}_{b}^{a} \\
        0 & 1
    \end{pmatrix},
\end{equation}
where ${\bf R}_{b}^{a} \in \mathbb{R}^{3 \times 3}$ is the rotation matrix from frame $a$ to frame $b$, and ${\bf p}_{b}^{a} \in \mathbb{R}^{3 \times 1}$ represents $b$'s origin expressed in frame $a$. 

At each control cycle, the stylus pose ${\bf T}_{\mathcal{H}_e}^{\mathcal{H}_b}$ is read by the HD driver, and converted in the robot end-effector pose ${\bf T}_{\mathcal{R}_e}^{\mathcal{R}_b}$ with the following kinematic equations:
\begin{align}
{\bf R}_{\mathcal{R}_e}^{\mathcal{R}_b} &= {\bf R}_{\mathcal{H}_b}^{\mathcal{R}_b} {\bf R}_{\mathcal{H}_e}^{\mathcal{H}_b} {\bf R}_{\mathcal{R}_e}^{\mathcal{H}_e}, \\
{\bf p}_{\mathcal{R}_e}^{\mathcal{R}_b} &= {\bf p}_{\mathcal{R}_e}^{\mathcal{R}_b}(0) + {\bf R}_{\mathcal{H}_b}^{\mathcal{R}_b} \big({\bf p}_{\mathcal{H}_e}^{\mathcal{H}_b} - {\bf p}_{\mathcal{H}_e}^{\mathcal{H}_b}(0)\big),
\end{align}
where $\mathcal{H}_e$ is the haptic stylus frame, $\mathcal{R}_e$ is the robot end-effector frame, and ${\bf p}_{\mathcal{R}_e}^{\mathcal{R}_b}(0)$ and ${\bf p}_{\mathcal{H}_e}^{\mathcal{H}_b}(0)$ are the starting positions of the robot end-effector and of the HD stylus, respectively. From ${\bf T}_{\mathcal{R}_e}^{\mathcal{R}_b}$, the corresponding robot end-effector pose is obtained as:
\begin{equation}
    {\bf x}_{\mathcal{R}_e}^{\mathcal{R}_b} = \left[ {\bf p}_{\mathcal{R}_e}^{\mathcal{R}_b}; {\bf q}_{\mathcal{R}_e}^{\mathcal{R}_b} \right],
\end{equation}
where ${\bf q}_{\mathcal{R}_e}^{\mathcal{R}_b} \in \mathbb{H}_1$ is the quaternion associated with ${\bf R}_{\mathcal{R}_e}^{\mathcal{R}_b}$. 

Commanding the robot with ${\bf x}_{\mathcal{R}_e}^{\mathcal{R}_b}$ would imply transferring the inevitable tremors of human motion, further emphasized by the control frequency, which is much higher than the typical bandwidth of hand tremors. Therefore, ${\bf x}_{\mathcal{R}_e}^{\mathcal{R}_b}$ is filtered online by a moving average filter. The desired position ${\bf p}_h^{\mathcal{R}_b}(i)$, at the $i$-th control cycle, is obtained by averaging samples from a recent past within a window of fixed size $n$. To achieve $\mathcal{O}(1)$ computational complexity, we keep a cumulative sum ${\bf c}(i)$, which is updated at each control cycle:
\begin{equation}
    {\bf c}(i) = \begin{cases}
        {\bf c}(i-1) + {\bf p}_{\mathcal{R}_e}^{\mathcal{R}_b}(i), & \text{if}~i < n; \\
        {\bf c}(i-1) + {\bf p}_{\mathcal{R}_e}^{\mathcal{R}_b}(i) - {\bf p}_{\mathcal{R}_e}^{\mathcal{R}_b}(i - n), & \text{otherwise.}
    \end{cases}
\end{equation}
The desired end-effector position ${\bf p}_h^{\mathcal{R}_b}(i)$ is obtained as:
\begin{equation}
    {\bf p}_h^{\mathcal{R}_b}(i) = \frac{{\bf c}(i)}{m},
\end{equation}
where $m \leq n$ is the number of samples in the moving average window at the $i$-th control cycle.
The desired orientation ${\bf q}_h^{\mathcal{R}_b}(i)$ is obtained by filtering ${\bf q}_{\mathcal{R}_e}^{\mathcal{R}_b}(i)$ as done in \cite{markley_averaging_2007}.

\subsection{Robot interaction control}\label{sec:robot-interaction-control}
In this section, we first formalize the contact model between the environment and the robot (Section \ref{sec:admittance-model}), then we present the interaction forces limitation strategy (Section \ref{sec:interaction-force-limitation-strategy}).

\subsubsection{Admittance model}\label{sec:admittance-model}
Let ${\bf x}_d^{\mathcal{R}_b} = {\bf x}_h^{\mathcal{R}_b} = \left[ {\bf p}_h^{\mathcal{R}_b}; {\bf q}_h^{\mathcal{R}_b} \right]$ be the desired task and ${\bf x}_c^{\mathcal{R}_e} = \left[ {\bf p}_c^{\mathcal{R}_e}; {\bf q}_c^{\mathcal{R}_e}\right]$ be the robot-commanded pose in end-effector frame. For the sake of simplicity we take ${\bf q}_c^{\mathcal{R}_e} = {\bf q}_d^{\mathcal{R}_e}$. When the robot interacts with the environment, the relation between the contact forces ${\bf f}^{\mathcal{R}_e} \in \mathbb R^{3 \times 1}$ and the position error $\tilde{{\bf p}} = {\bf p}_d^{\mathcal{R}_e} - {\bf p}_c^{\mathcal{R}_e}$ is established by a generalized mechanical impedance, represented by a second-order mass-spring-damper system:
\begin{equation}\label{eq:admittance-model}
    {\bf M}_d \ddot{\tilde{{\bf p}}}^{\mathcal{R}_e} + {\bf K_D} \dot{\tilde{{\bf p}}}^{\mathcal{R}_e} + {\bf K}_P \tilde{{\bf p}}^{\mathcal{R}_e} = - \tilde{{\bf f}}^{\mathcal{R}_e},
\end{equation}
where ${\bf M}_d, {\bf K}_D, {\bf K}_P$ are the mass, damping, and stiffness diagonal matrices, respectively, of proper size, which are used to impose specific dynamics. 
The force tracking error $\tilde{{\bf f}}^{\mathcal{R}_e} = {\bf f}_d^{\mathcal{R}_e} - {\bf f}_e^{\mathcal{R}_e}$ is the difference between the desired force ${\bf f}_d^{\mathcal{R}_e}$ and the exerted force ${\bf f}_e^{\mathcal{R}_e}$, which is equal and opposite to the measured force. 

Computing ${\bf p}_c^{\mathcal{R}_e}$ from \eqref{eq:admittance-model}, the compliant position reference to be commanded to the robot is obtained as:
\begin{equation}
    {\bf p}_c^{\mathcal{R}_b} = {\bf p}_d^{\mathcal{R}_b} + {\bf R}_{\mathcal{R}_e}^{\mathcal{R}_b}{\bf p}_c^{\mathcal{R}_e}.
\end{equation}

\subsubsection{Interaction force limitation strategy}\label{sec:interaction-force-limitation-strategy}
Let ${\bf f}_d^{\mathcal{R}_e} = {\bf 0}$ to make the robot compliant with the environment and assume that the latter can be approximated by a linear spring model. The exerted force at the steady state \cite{gao_combined_2023} is:
\begin{equation} \label{eq:desired-position-for-force-reference-tracking}
    {\bf f}_e^{\mathcal{R}_e} = {\bf K}_{eq} ({\bf p}_e^{\mathcal{R}_e} - {\bf p}_d^{\mathcal{R}_e}),
\end{equation}
where ${\bf K}_{eq} = ({\bf K}_P {\bf K}_e)({\bf K}_P + {\bf K}_e)^{-1}$ is the equivalent stiffness of the admittance stiffness ${\bf K}_P$ and environmental stiffness ${\bf K}_e$ \cite{li_fuzzy_2021}, ${\bf p}_e^{\mathcal{R}_e}$ is the environment rest position. From \eqref{eq:desired-position-for-force-reference-tracking}, it follows that contact forces can be bounded through a suitable reference ${\bf p}_d^{\mathcal{R}_e}$ provided to the admittance controller. Accurate limitation requires perfect knowledge of the environment.

In the proposed strategy, the $j$-th component of the position reference is saturated when the corresponding component of the contact force exceeds a limit, i.e.,
\begin{equation} \label{eq:saturation}
    p_{d, j}^{\mathcal{R}_b} = 
    \begin{cases}
    p_{h, j}^{\mathcal{R}_b}, &\quad \text{if } |f_{e, j}^{\mathcal{R}_b}| < f_{th, j}\\
    \overline{p}_{d, j}^{\mathcal{R}_b}, &\quad \text{if } |f_{e, j}^{\mathcal{R}_b}| > f_{th, j}
    \end{cases},
\end{equation}
where $ f_{th, j} \in \mathbb{R}^+$ and $\overline{p}_{d, j}^{\mathcal{R}_b}$, $\forall j \in \left\{ x, y, z \right\}$, are the saturation activation threshold and the reference saturation value along each axis of the base reference frame, respectively. From \eqref{eq:desired-position-for-force-reference-tracking}, we set
\begin{equation}
    \overline{p}_{d, j}^{\mathcal{R}_b} = p_{e, j}^{\mathcal{R}_b} + \frac{f_{th, j}}{K_{eq, j}},
\end{equation}
where ${\bf p}_e^{\mathcal{R}_b}$ can be obtained through sensor sampling. In our design, assuming a stiff environment with uniform geometry, we approximate it by considering the last compliant position commanded to the robot ${\bf p}_c^{\mathcal{R}_b}$ at the saturation activation. With regard to the environment stiffness, we assume that ${\bf K}_e$ has been previously estimated, e.g., as in \cite{roveda_sensorless_2022}, and it does not change over time. We envisage the possibility of not having an accurate estimate, but rather a conservative one that allows for an effective forces limitation. Likewise, we note that $f_{th, j}$ does not represent an accurate force limit, as it is always exceeded before the saturation mechanism can take place. However, the strategy in \eqref{eq:saturation} prevents that contact forces can grow indefinitely. The force threshold can be devised from the task requirements, or, if available, from observations/measurements of human operators manually performing the same task. 

\begin{figure}
    \centering
    \subfloat[No contact state. \label{fig:}]{\includegraphics[width=0.498\columnwidth]{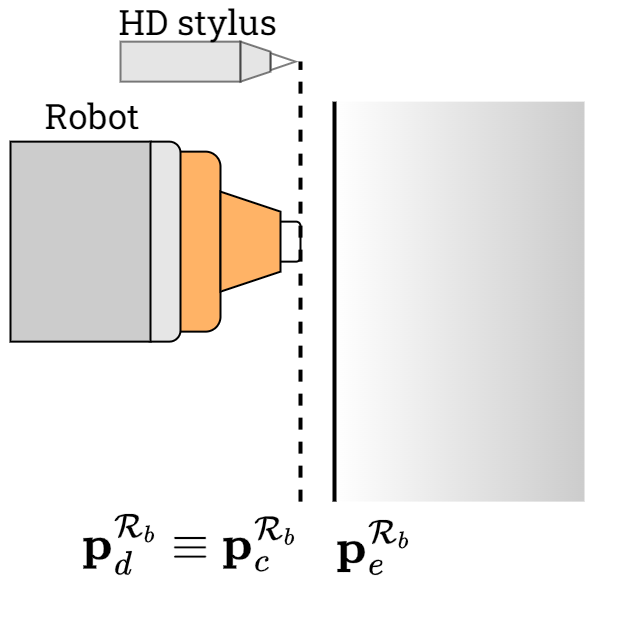}}
    \hfill
    \subfloat[Collision state. \label{fig:}]{\includegraphics[width=0.498\columnwidth]{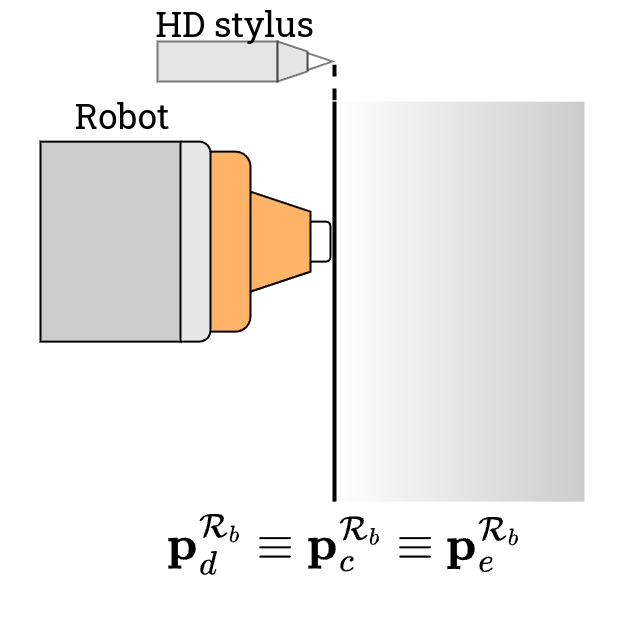}}
    \\
    \subfloat[Penetration state. \label{fig:scenario-b-outcome}]{\includegraphics[width=0.498\columnwidth]{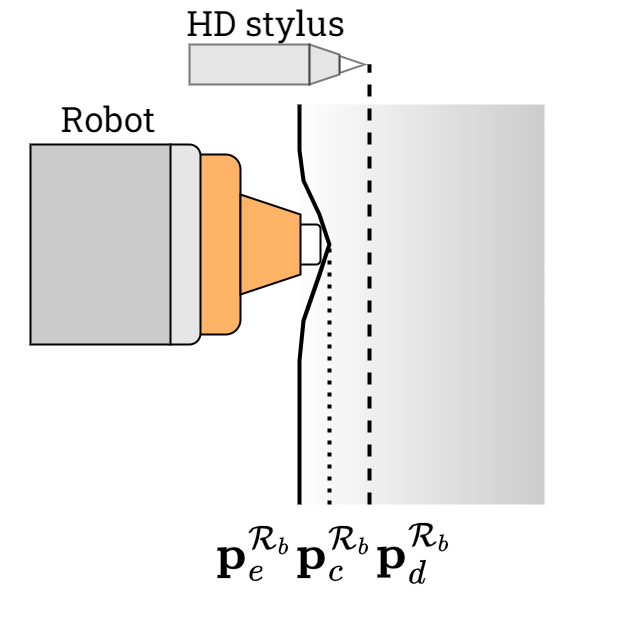}}
    \hfill
     \subfloat[Saturation state. \label{fig:scenario-c-outcome}]{\includegraphics[width=0.498\columnwidth]{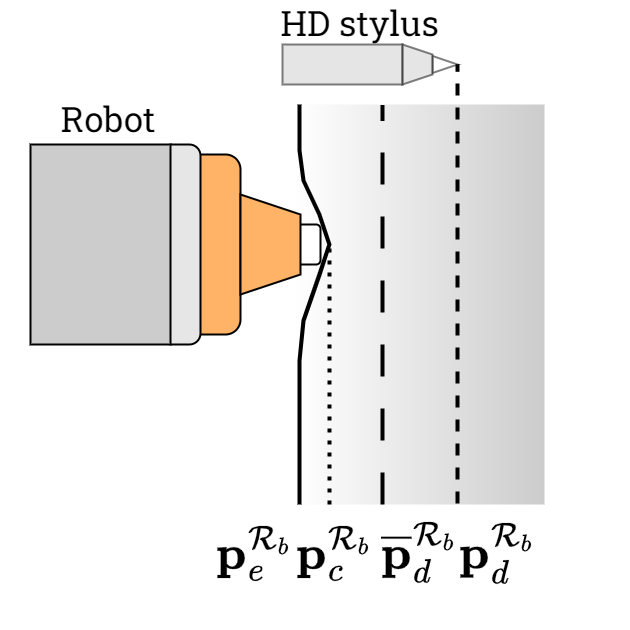}}
    \caption{Operation of the robot interaction control in the different contact states with the environment.}
    \label{fig:interaction-control-states}
\end{figure}

A pictorial view of the robot-environment contact phases is provided in Figure \ref{fig:interaction-control-states}, with an indication of the symbols used in this paper.

\subsection{Force feedback rendering}\label{sec:force-feedback-rendering}
In impedance systems, such as the mass-spring-damper in \eqref{eq:admittance-model}, interactions are captured by both ${\bf f}_e^{\mathcal{R}_e}$ and $\tilde{\bf p}^{\mathcal{R}_e}$. The former exhibits step dynamics, while the latter, being the result of a second-order system, is characterized by slower transients. Displaying ${\bf f}_e^{\mathcal{R}_e}$ directly would make the transient of the interaction event unperceivable for the human operator, who would have to cope with sudden force changes they are not used to from their previous experience of interacting with the environment. Therefore, in our rendering strategy, we replace the direct measurements ${\bf f}_e^{\mathcal{R}_e}$ with $\tilde{\bf p}^{\mathcal{R}_e}$, and define a second impedance model generating the haptic forces:
\begin{equation} \label{eq:force-feedback}
    {\bf f}_h^{\mathcal{H}_b} = {\bf K}_{h} \tilde{{\bf p}}_h^{\mathcal{H}_b} + {\bf D}_{h} \dot{\tilde{{\bf p}}}_h^{\mathcal{H}_b},
\end{equation}
where ${\bf K}_{h}$ and ${\bf D}_{h}$ respectively are stiffness and damping diagonal matrices of the proper size,
\begin{equation}
    \tilde{{\bf p}}_h^{\mathcal{H}_b} = {\bf R}_{\mathcal{R}_b}^{\mathcal{H}_b} \tilde{{\bf p}}^{\mathcal{R}_b} = {\bf R}_{\mathcal{R}_b}^{\mathcal{H}_b} \left({\bf p}_h^{\mathcal{R}_b} - {\bf p}_c^{\mathcal{R}_b}\right)
\end{equation}
is the unsaturated position error, and
\begin{equation}
    \dot{\tilde{{\bf p}}}_h^{\mathcal{H}_b} =  {\bf R}_{\mathcal{R}_b}^{\mathcal{H}_b} {\bf R}_{\mathcal{R}_e}^{\mathcal{R}_b} \dot{\tilde{{\bf p}}}^{\mathcal{R}_e}
\end{equation}
its time derivative. The choice of using the unsaturated position error is motivated by the interest of not limiting the haptic feedback and exploring the full range of forces that can be exerted by the HD.

Indeed, the system in \eqref{eq:force-feedback} takes the role of virtualizing the contact force and decoupling it from the admittance control dynamics. The parameters ${\bf K}_h$ and ${\bf D}_h$ are tuned to provide a sense of naturalness while the SR interacts with the environment, regardless of the actual degree of compliance employed by the onboard admittance controller. In particular, ${\bf K}_h$ can be tuned to ensure a desired transient response, while ${\bf D}_h$ can be tuned to reduce the vibrations that occur especially when the robot slips along the tangential directions.

\section{Experiments}
\begin{figure}
\centering
\includegraphics[width=\columnwidth]{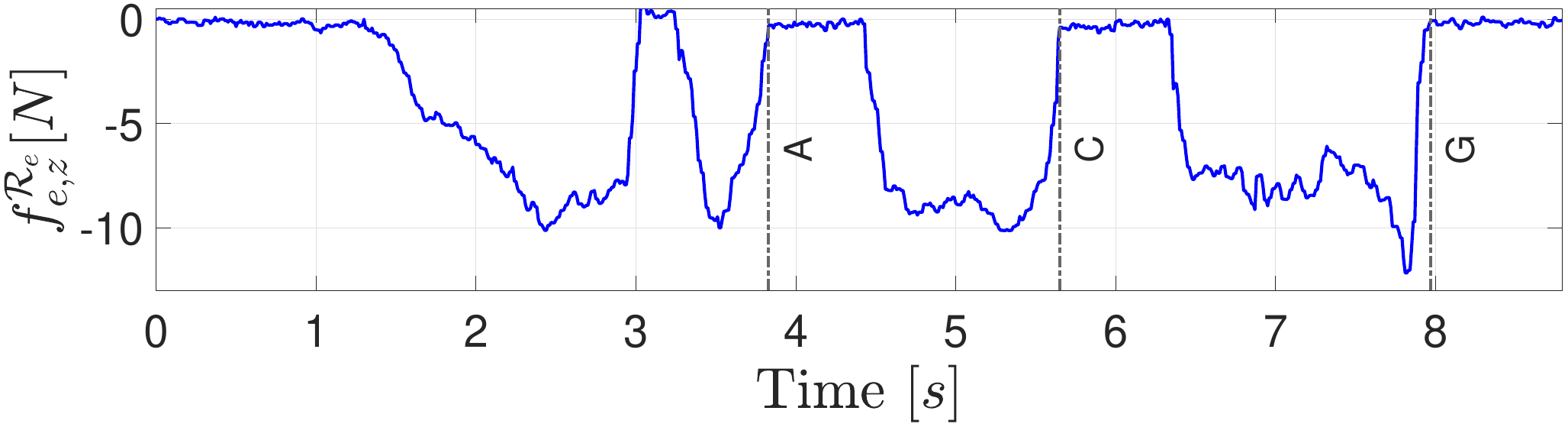}
\caption{Human force profile representative of freehand writing for the sequence \emph{A-C-G}.}
\label{fig:human-force-profiles}
\end{figure}

\begin{figure}
\centering
\includegraphics[width=0.85\columnwidth, height=3.8cm]{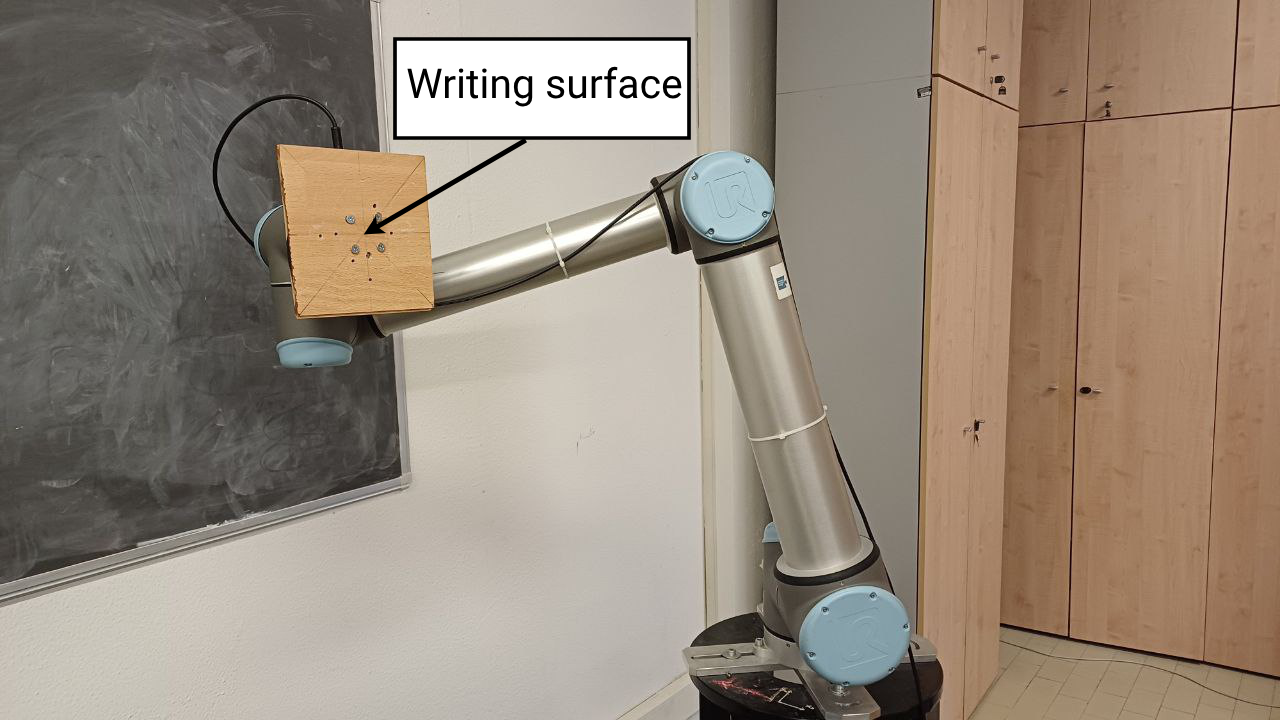}
\caption{Experimental setup to record human force profiles.}
\label{fig:freehand-writing-setup}
\end{figure}

To validate the performance of the proposed HBTS and perform a quantitative assessment of different aspects of the system, we conduct blackboard writing experiments with chalk, using three different architectures. The chosen task is as simple as it is significative, since it requires a naturally-rendered force feedback to achieve accurate handwriting, as well as a precise control of interaction to ensure the integrity of the chalk, both of which are crucial requirements for, e.g., medical procedures. With the aim of confirming the impact of the proposed HBTS on the naturalness of teleoperation and the safety of interaction, we compare the teleoperation experiments with manual handwriting, in terms of attained force profiles. Figure \ref{fig:human-force-profiles} shows a typical force profile of a freehand writing task, corresponding to the sequence of letters \textit{A-C-G} (\emph{Automatic Control Group}), recorded on a small force-sensed surface, oriented like a blackboard on the wall, as shown in Figure \ref{fig:freehand-writing-setup}.

In order to compare the performance of the different architectures, we define three evaluation metrics, one qualitative and two quantitative ones:
accuracy and continuity of the handwriting and similarity with human force profiles, in terms of the mean difference (MD) of the forces exerted by the human and robot and of the absolute difference between their maximum values. The MD is computed as:
\begin{equation}
    MD = \left| \frac{1}{N} \sum_{i=1}^N {}^{hum}f_{e, z}^{\mathcal{R}_e}(i) - \frac{1}{M} \sum_{i=1}^M {}^{rob}f_{e, z}^{\mathcal{R}_e}(i)\right|,
\end{equation}
where $M$ and $N$ are the number of samples collected by the human and robot execution, respectively. The mean value of the forces exerted by the human is $\mu_{hum} = -7.25\,N$. The absolute difference between their maximum values is computed as:
\begin{equation}
   \Delta\overline{f}_{{e, z}}^{\mathcal{R}_e} = \left|{}^{hum}\overline{f}_{{e, z}}^{\mathcal{R}_e} - {}^{rob}\overline{f}_{{e, z}}^{\mathcal{R}_e}\right|,
\end{equation}
where ${}^{hum}\overline{f}_{{e, z}}^{\mathcal{R}_e} = -12.1600\,N$.
In Section \ref{sec:ft-measure-based-rendering}, we report the results of the task execution through a HBTS, featuring a measurement-based force feedback (Scenario A). In such architecture the saturation of the admittance controller's motion references is disabled, therefore references beyond the maximum penetration are accepted and commanded.
In Section \ref{sec:motion-error-based-rendering-no-saturation}, we report the results of the task execution through a HBTS, featuring the motion-error-based force feedback (Scenario B). As in the previous case, the references saturation is disabled.
Finally, in Section \ref{sec:motion-error-based-rendering-saturation}, we report the results of the task execution through the HBTS proposed in this paper (Scenario C).

\begin{figure}
    \centering
    \subfloat[Scenario A. \label{fig:scenario-a-outcome}]{\includegraphics[width=0.325\columnwidth]{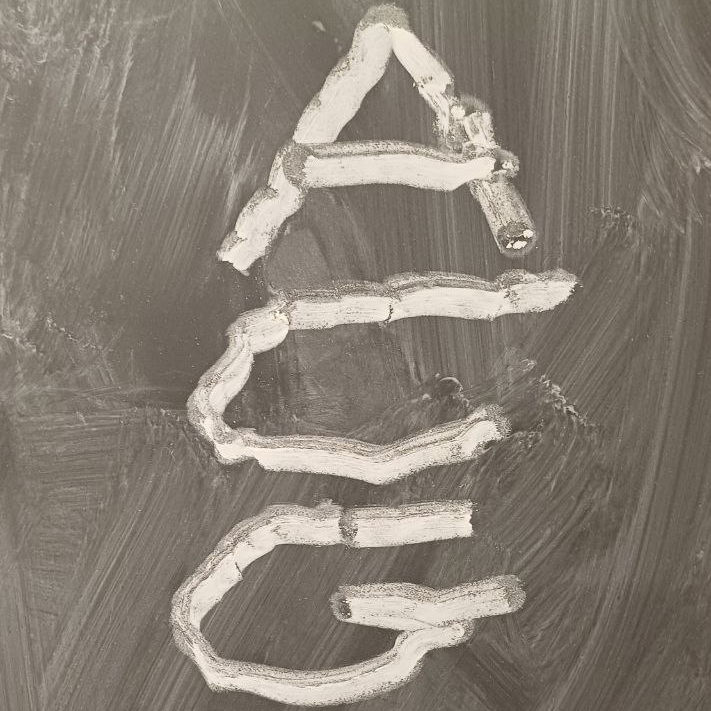}}
    \hfill
    \subfloat[Scenario B. \label{fig:scenario-b-outcome}]{\includegraphics[width=0.325\columnwidth]{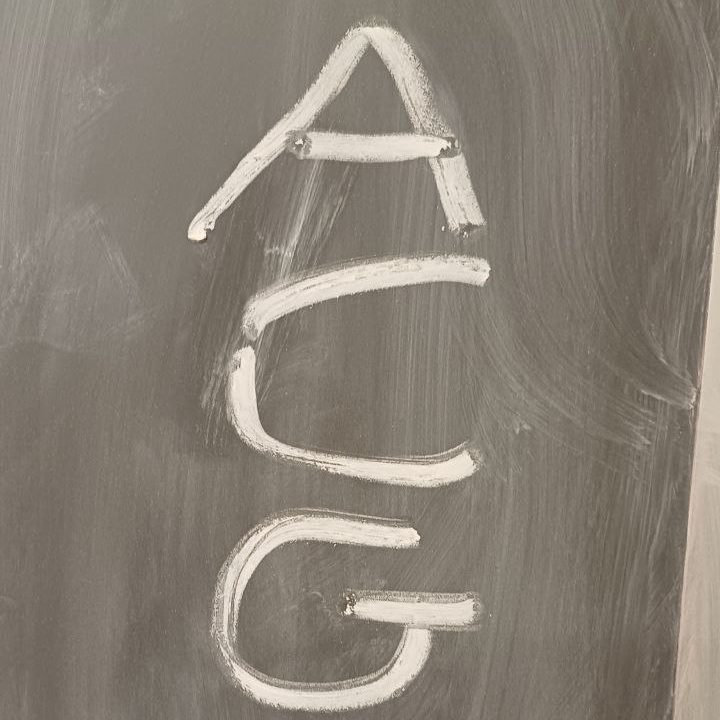}}
    \hfill
     \subfloat[Scenario C. \label{fig:scenario-c-outcome}]{\includegraphics[width=0.325\columnwidth]{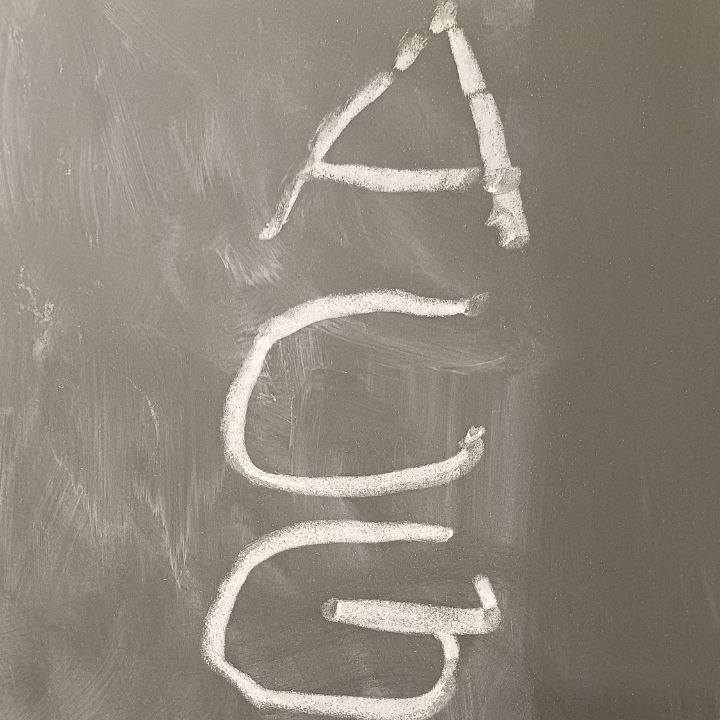}}
    \caption{Outcomes of the blackboard writing task execution in the three scenarios.}
    \label{fig:experiment-outcomes}
\end{figure}

\subsection{Force/torque measure-based force feedback} \label{sec:ft-measure-based-rendering}
In order to assess the improved naturalness of the proposed virtualized force feedback, we execute the task with the typical rendering strategy directly from the force sensed at the robot's end-effector. We adopt a non-linear mapping, based on hyperbolic cosine, to map such forces, in the order of tens of N, to the HD, so that they are in the order of units, and have a slower growth rate than linear mapping. The chosen mapping is:
\begin{equation}\label{eq:hyperbolic-cosine-mapping}
    f_{h, j}^{\mathcal{H}_b} = \cosh{\left(\frac{1}{\overline{f}_{{e, j}}^{\mathcal{H}_b}} \ln{\left(4 + \sqrt{15}\right)} f_{e, j}^{\mathcal{H}_b}\right)},
\end{equation}
where
\begin{equation}
    {\bf f}_e^{\mathcal{H}_b} = {\bf R}_{\mathcal{R}_b}^{\mathcal{H}_b} {\bf R}_{\mathcal{R}_e}^{\mathcal{R}_b}{\bf f}_e^{\mathcal{R}_e},
\end{equation}
and
\begin{equation}
     \overline{{\bf f}}_e^{\mathcal{H}_b} = {\bf R}_{\mathcal{R}_b}^{\mathcal{H}_b} {\bf R}_{\mathcal{R}_e}^{\mathcal{R}_b} \overline{{\bf f}}_e^{\mathcal{R}_e},
\end{equation}
with $\overline{{\bf f}}_e^{\mathcal{R}_e}$ being the maximum expected contact force along each axis of the end-effector frame. For the task execution, we set the value of each component to $12\,N$.

We execute the task $3$ times before obtaining a successful execution. The first $2$ failures are due to the forces exerted exceeding the safety threshold, beyond which the robot stops and the chalk breaks or shatters. 

\begin{figure}
\centering
\includegraphics[width=\columnwidth]{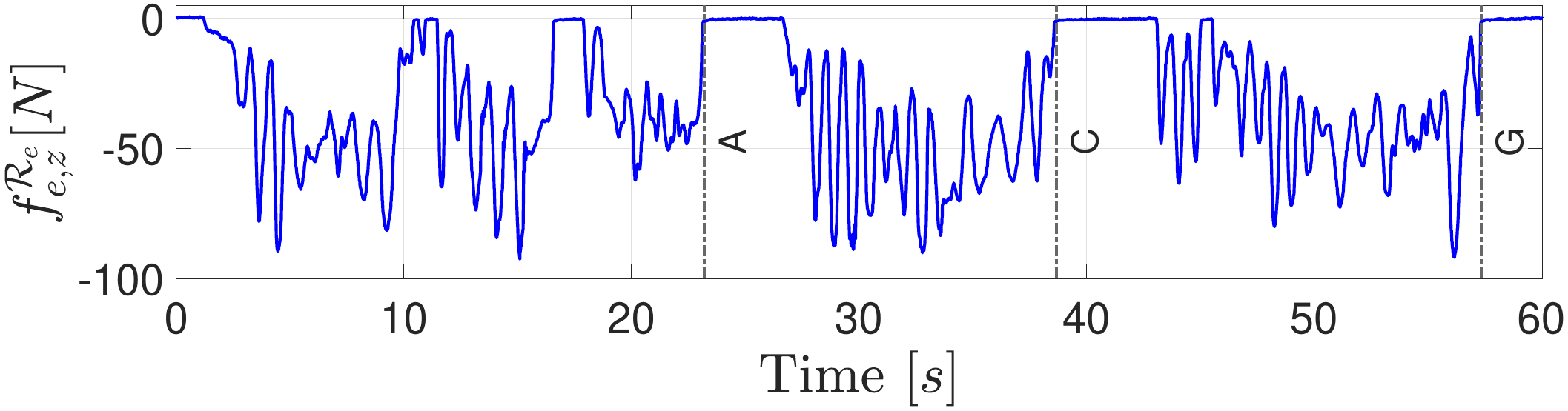}
\caption{Interaction forces along the axis perpendicular to the blackboard plane in Scenario A.}
\label{fig:scenario-a-interaction-forces}
\end{figure}

The outcome of the successful execution of the task is shown in Figure \ref{fig:scenario-a-outcome}. Visual inspection of the result showcases poor handwriting accuracy, with evident stroke discontinuities. This is due to the sudden changes in force feedback that cause strong backward pushes of the HD, resulting in a deterioration of writing performance.
Regarding the interaction forces, the overall results, shown in Figure \ref{fig:scenario-a-interaction-forces}, highlight a highly oscillatory behavior, with $MD = 48.32\,N$ and $\Delta \overline{f}_{{e, z}}^{\mathcal{R}_e} = 80.40\,N$. 

\subsection{Virtualized force feedback without references saturation} \label{sec:motion-error-based-rendering-no-saturation}
Since we aim to overcome the limitations of measurement-based force feedback, we execute the task with the proposed virtualized force feedback. As for scenario A, the saturation of the admittance reference is disabled.

We execute the task $2$ times before obtaining a successful execution. Again, the first failure is due to the forces exerted exceeding the safety threshold.

\begin{figure}
\centering
\includegraphics[width=\columnwidth]{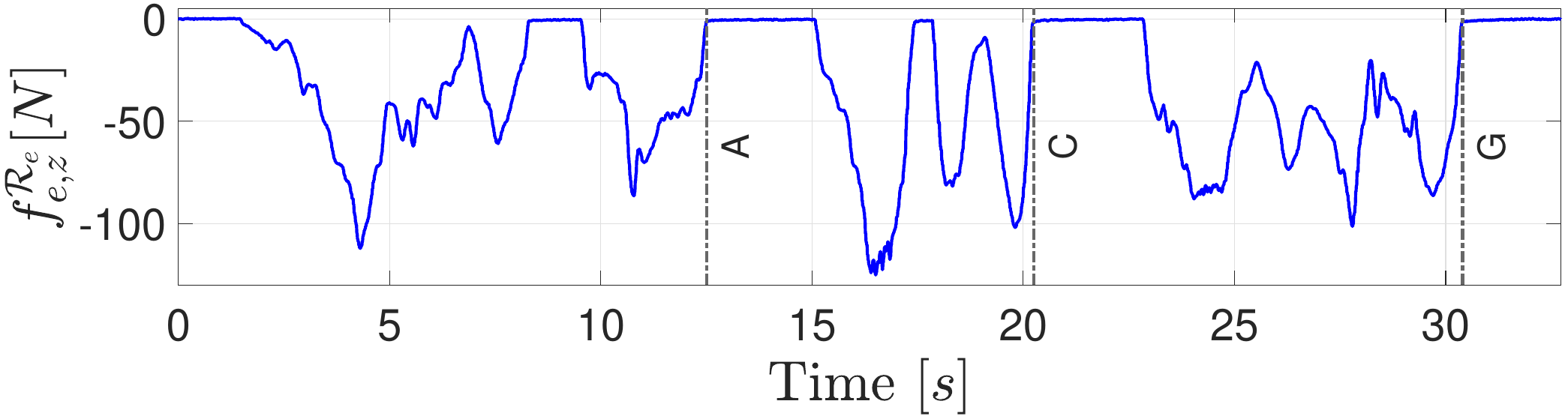}
\caption{Interaction forces along the axis perpendicular to the blackboard plane in Scenario B.}
\label{fig:scenario-b-interaction-forces}
\end{figure}

The outcome of the successful task execution is shown in Figure \ref{fig:scenario-b-interaction-forces}. Visual inspection of the result showcases a significant improvement in writing accuracy: indeed, the letters show an ideal height-to-width ratio \cite{somberg_character_1990}, with significantly improved stroke continuity.
Regarding the interaction forces, the overall results, shown in Figure \ref{fig:scenario-b-outcome}, do not guarantee an improvement over Scenario A; on the contrary, $MD = 54.88\,N$ and $\Delta \overline{f}_{e, z}^{\mathcal{R}_e} = 137.21\,N$. 

\subsection{Virtualized force feedback with references saturation} \label{sec:motion-error-based-rendering-saturation}

In the last scenario, we execute the task by foreseeing both the virtualized force feedback rendering methodology and the saturation of the motion references provided to the admittance controller. For the execution of the procedure we define the saturation activation threshold along the axis perpendicular to the interaction plane to be $f_{th,x} = 12\,N$, corresponding to the maximum force exerted by the human operator in the freehand execution.

\begin{figure}
\centering
\includegraphics[width=\columnwidth]{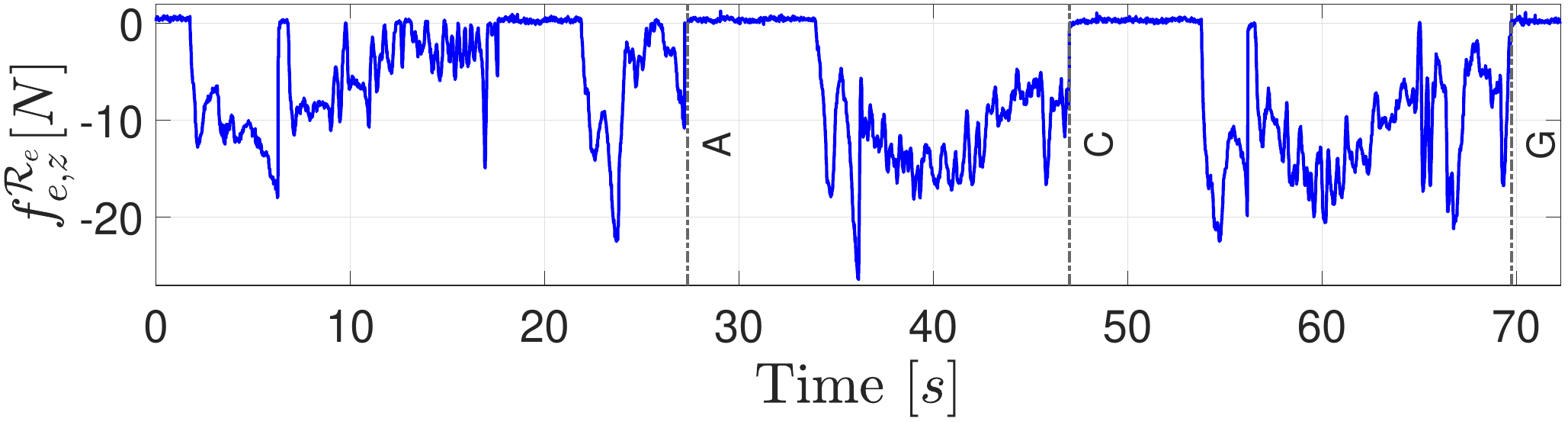}
\caption{Interaction forces along the axis perpendicular to the blackboard plane in Scenario C.}
\label{fig:scenario-c-interaction-forces}
\end{figure}

The task is successful on the first trial, whose outcome is shown in Figure \ref{fig:scenario-c-outcome}. As expected, the results are consistent with what was seen in the previous scenario. However, from Figure \ref{fig:scenario-c-interaction-forces}, we observe a significant reduction in contact forces, with $MD = 1.40\,N$ and $\Delta \overline{f}_{{e, z}}^{\mathcal{R}_e} = 14.22\,N$.  

\begin{figure}
\centering
\includegraphics[width=\columnwidth]{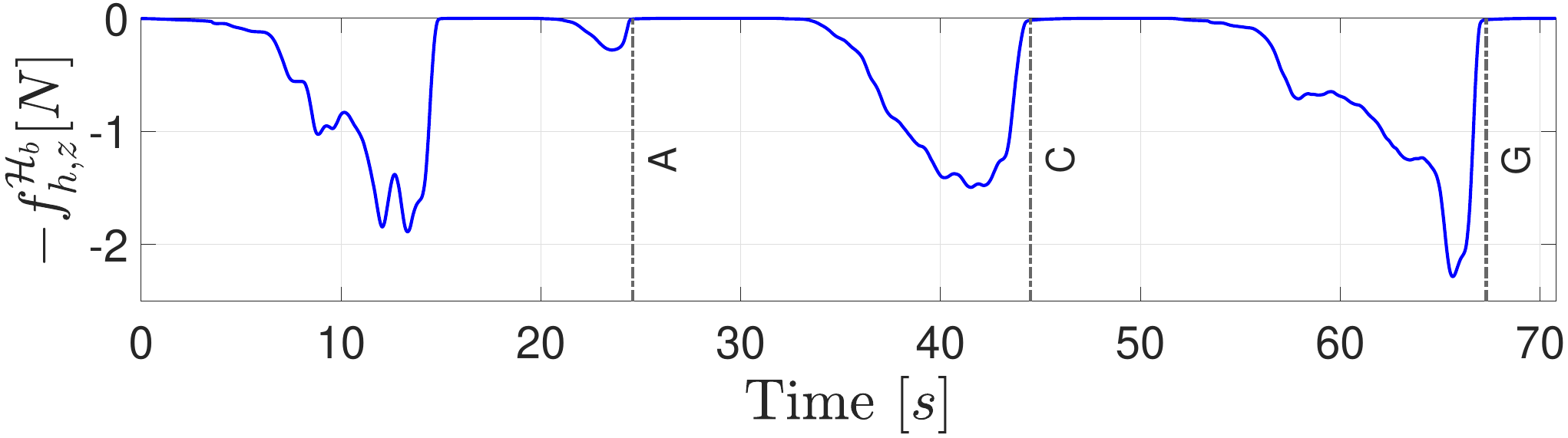}
\caption{Forces rendered on the HD in Scenario C.}
\label{fig:haptic-force-profiles}
\end{figure}

The impact of the architecture on the naturalness of execution can be captured from Figure \ref{fig:haptic-force-profiles}, where the forces rendered on the HD in Scenario C are shown. Visual inspection of the result showcases a high similarity to the force profiles obtained from freehand writing, as well as slower dynamics than contact forces. This aspect improves both naturalness and comfort for the human operator.

\section{Conclusions and future works}
This paper proposed a new HBTS to facilitate the execution of precision tasks using haptic teleoperations. We proposed a new virtualized force feedback that enables a natural perception of contact states, increasing human operator awareness and system stability. By saturating the motion references provided to an admittance-controlled robot, the safety of the interaction was increased, with the possibility of fulfilling more stringent requirements in relation to the current certification process for pHRI applications. 
Experiments illustrated that the proposed architecture improves both the accuracy and safety of the task, preventing damage to the instrumentation, or the manipulated environment.  

The conducted trials showed that the reference saturation strategy offers significant advantages in terms of limiting contact forces during the execution of a precision task, but gives no guarantees of a strict limitation below a desired bound, mainly because of the approximate knowledge of the environment. This is particularly evident in contact transition states, where saturation mechanisms are not active yet. 

Based on the experimental findings of this paper, with our future work we will: (1) assess the employment of variable impedance schemes, as well as methodologies for estimating the environment properties, to limit the forces in all contact states; (2) rigorously study and ensure the stability of the HBTS, even in the presence of communication delays and packet loss, extending the proposed architecture with master-slave control strategies based on passivity \cite{sheng_time_2019}; (3) derive autonomous data-driven control policies \cite{zhang_learning_2021} exploiting haptic teleoperation data, since the poses commanded on the HD, together with resulting forces, can be interpreted as an expert policy; (4) validate the proposed HBTS in a medical use case, where both naturalness and safety are crucial.

\bibliographystyle{IEEEtran}
\bibliography{case-2024-haptic-teleop}

\end{document}